\documentclass[letterpaper,10pt,conference,twocolumn,romanappendices]{ieeeconf2}

\IEEEoverridecommandlockouts
\usepackage{amsmath,amssymb,theorem}
\usepackage{graphicx}
\graphicspath{ {./images/} }
\usepackage{algorithm,algorithmic}
\usepackage{psfrag}
\usepackage{cite}

\usepackage{color,xspace}
\usepackage{subfigure}
\usepackage{microtype}
\newtheorem{theorem}{Theorem}[section]

\newtheorem{problem}[theorem]{Problem}

\newcommand{\real}{{\mathbb{R}}}

\newcommand{\UU}{{\mathcal{U}}}

\newcommand{\timestep}{\Delta t}

\renewcommand{\epsilon}{\varepsilon}

\newcommand{\until}[1]{\{1,\dots, #1\}}

\newcommand{\oprocendsymbol}{\hbox{$\bullet$}}
\newcommand{\oprocend}{\relax\ifmmode\else\unskip\hfill\fi\oprocendsymbol}

\renewcommand{\theenumi}{(\arabic{enumi})}

\begin{document}

\title{ Simulate Less, Expect More: Bringing Robot Swarms to Life via Low-Fidelity Simulations}

\author{Ricardo Vega\qquad Kevin Zhu \qquad Sean Luke \qquad Maryam Parsa \qquad Cameron Nowzari \thanks{The authors are
    with the Department of Electrical and Computer Engineering and Department of Computer Science,
    George Mason University, Fairfax, VA 22030, USA, {\tt\small
      \{rvega7,kzhu4,sean,cnowzari,mparsa\}@gmu.edu}}}

\maketitle

\begin{abstract}
This paper proposes a novel methodology for addressing the simulation-reality gap for multi-robot swarm systems. Rather than immediately try to shrink or `bridge the gap' anytime a real-world experiment failed that worked in simulation, we characterize conditions under which this is actually necessary. When these conditions are not satisfied, we show how very simple simulators can still be used to both (i) design new multi-robot systems, and (ii) guide real-world swarming experiments towards certain emergent behaviors when the gap is very large. The key ideas are an iterative simulator-in-the-design-loop in which real-world experiments, simulator modifications, and simulated experiments are intimately coupled in a way that minds the gap without needing to shrink it, as well as the use of minimally viable phase diagrams to guide real world experiments. We demonstrate the usefulness of our methods on deploying a real multi-robot swarm system to successfully exhibit an emergent milling behavior. 
\end{abstract}

%\keywords{Swarm Robotics, Emergent Behaviors, Simulation-Reality Gap, Multi-Agent System}

\section{Introduction}
Swarms of robots have been theorized to help in all sorts of practical problems including search and rescue \cite{RA-JJ-BA-EM:20}, pollution monitoring \cite{GZ-GKF-DPG:11}, surveillance \cite{MS-JC-LP-JT-GL-AT-VV-VK:14}, and disaster management systems \cite{HK-CWF-IT-BS-ER-KP-AW-JW-12}; however, despite so much research they have not yet found much practical use. The potential benefits are clear as shown in many different simulations~\cite{JG-SMO-ALC:18,SAE-JM-OY-KG:18,AL-MO-NW:22}, but recreating these simulated behaviors on real robotic swarms is not a trivial extension due to physical limitations on actuation/sensing, imperfect communication, or collisions that are often overlooked or oversimplified in the simulated world. Applying these algorithms to real robots can often result in severe issues or failure not observed in simulation~\cite{ HH-TA-AR:20, GV-CV-GS-TN-AEE-TV:18}. The issues with the simulation-reality gap and the desire to bridge it is not new~\cite{NJ-PH-IH:95, SK-JM-SD:10, SK-JM-SD:12, JT-MR-NY-SC-DB-AR:22}. However, unlike the majority of works that discuss this for simulators of a single robot system, we propose a novel approach that addresses the simulation-reality gap without necessarily attempting to shrink this gap unless deemed necessary. 

Specifically, this work aims to achieve two major goals. First, we want to leverage decades worth of research in discovering different swarming/emergent behaviors that have only been demonstrated in simulation~(e.g., Artificial Life) by showing how these agent-based modeling tools can be useful for both designing robot swarms and guiding experiments in real time. Second, we want to provide a framework to allow a robot swarm engineer to systematically determine whether a team of available robots is actually able to recreate some behavior found in simulation, and how exactly to make it happen.

\begin{figure}[h]
    \centering
    \subfigure[]{\includegraphics[width=.35\linewidth]{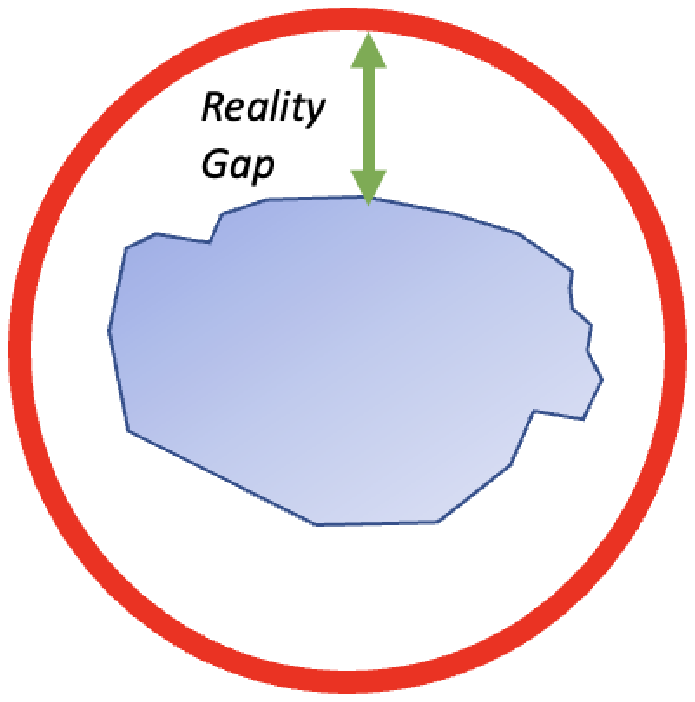}}
    \subfigure[]{\includegraphics[width=.35\linewidth]{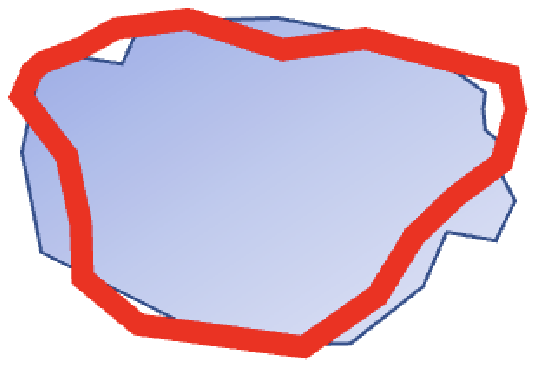}}
    \subfigure[]{\includegraphics[width=.35\linewidth]{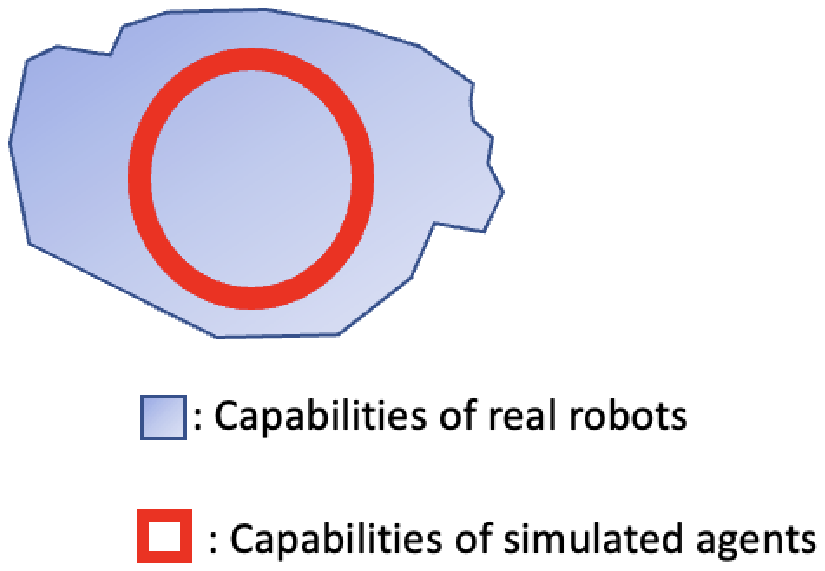}}
    \caption{Cartoon showing the capabilities of robots (with respect to their environments) in simulation (in red) versus reality (in blue) in (a) traditional robot swarm simulation research, (b) the case of a high-fidelity simulator paired with a matching multi-robot platform, and (c) our approach of making our simulated agents less capable than their real-world counterparts.}
    \label{fig:gap}
\end{figure}

To solve this problem we kept two things in mind. Simulations should somehow be representative of real-world experiments but also be simple enough to enable rapid modification, development, and utilization. The traditional approach to creating a simple simulator of a real-world robotic system is shown in Fig.~\ref{fig:gap}(a), where the simulated agent often starts in an idealized disturbance/noise-free environment with perfect sensing, actuation, and communication abilities (shown as a red circle). This proof-of-concept simulation is then used as a benchmark or goal to build a robot with capabilities as close as possible to the simulation (shown as a smaller and rougher blue shape). The experiment can then be done with the real robot and everyone can hope it works as close enough to the simulator as needed. Naturally a physical robot will not be able to keep up with a perfectly simulated agent in a disturbance-free world, but that doesn't necessarily mean the experiment will fail. When the experiment fails and the real robot is not able to replicate what is observed in simulation, the traditional approach is to shrink the gap in one of two ways: either (i) begin simulating more real-world considerations (e.g., characteristics of sensors and actuators, environmental conditions, or friction) and shrinking the capabilities of the simulated agent; or (ii) upgrade the robot (in hardware and/or software) to expand the capability of the physical robot. In either case the goal is to bring the capabilities closer. Fig.~\ref{fig:gap}(b) then shows a higher fidelity simulator for a given system achieved after spending time painstakingly programming the different actual disturbances into a high fidelity simulator. It should be acknowledged that for a dedicated robot for which design changes are minimal, it is perfectly natural to take the time to build a high fidelity simulator of the real robot system as well as possible. However, for doing research in robotic swarm systems, the time to do this is prohibitive as even minor changes/adjustments made to the team of robots may require a lot of time to update in the simulator and validate again. 

The main idea of our alternate approach is shown in Fig.~\ref{fig:gap}(c) where we are purposely making our simulated robot \textit{less} capable than physically existing robots in every aspect possible. The idea behind this is if we make the simulated robot's capabilities worse and its environment harsher, demonstrating success in simulation may serve as a sufficient condition for the real-world experiment working rather than simply a (often unattainable) design goal.

A key aspect of our approach is that we fully acknowledge the fact that, although the multi-agent system of robots may be meant to be homogenous, there most certainly are differences from robot to robot (or idiosyncrasies). However, researchers in sociobiology studying insects have found that temperatures in bumble bee nests are able to reach safe stable values due to the different thresholds of the individual bees; some bees would start fanning (cooling) or incubating (heating) at different temperatures than others and these idiosyncrasies allow for a stable homeostatic state~\cite{SO-RLF:01}. So rather than try to diminish or eliminate these idiosyncrasies, we maintain these differences and attempt to use them to our advantage in our goal to deploy real swarming robots more reliably and predictably than is done today.

\begin{figure*}[ht]
\centering
    \includegraphics[width=.8\linewidth]{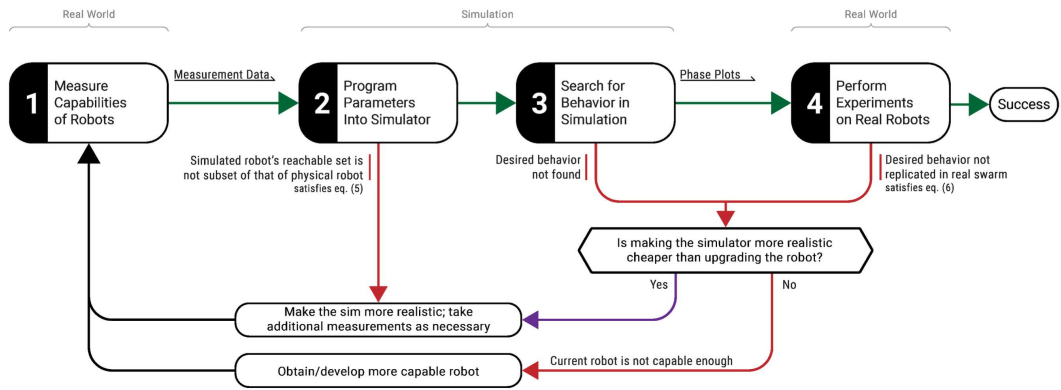}
    \caption{Flow Chart of RSRS Process}\label{fig:steps}
\end{figure*}

\paragraph*{Statement of Contributions} 
In order to realize the two major goals described above, our first contribution is this novel connection between swarm simulator and physical robot platform in which the simulated world is harsher than reality (Steps 1-3 in Fig.~\ref{fig:steps} connecting the real world to the simulated world). Our second contribution is the method of using minimally viable phase diagrams to make informed decisions on whether available hardware is sufficient in demonstrating a particular swarming behavior or some form of upgrade is necessary (e.g., a better sensor or a faster controller). Inspired by phase diagrams from chemistry~\cite{BP-MH-MJP:13} that taught us in high school how to create a swarm of water molecules to exhibit different global properties (e.g., liquid, solid, or gas) by controlling temperature and pressure, we wish to create and immediately make use of similar diagrams for better controlling robot swarms. Researchers in simulation have been able to create these phase diagrams for different multi-agent systems \cite{AC-CKH:18, MRD-YC-ALB-LSC:06, NVB-HA-IYT-SAM:20, ZC-ZC-VT-DC-HZ:16}, albeit using capable individual agents that require a lot of information like the position and/or the orientation of it's neighbors. Our proposed framework is applied to three different real multi-robot systems to recreate a milling behavior using a simple direct sensing-to-action controller using a binary sensor.

\section{Problem Formulation}\label{se:pre}

\begin{figure}[h]
    \centering
    \includegraphics[width=4cm]{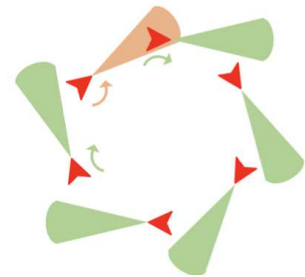}
    \caption{Milling circle formed using a simple binary sensing-to-action controller: if something is detected steer left, otherwise steer right~\cite{FB-MG-RN:21,MG-JC-TJD-RG:14}.}
    \label{fig:milling_figure}
\end{figure}

In this paper we consider only a particular emergent \textit{milling} behavior that has already been demonstrated and recreated in multiple different simulators and research labs around the world \cite{AC-CKH:18, MG-JC-TJD-RG:14, DB-RT-OH-SL:18, DS-CP-GB:18,FB-MG-RN:21}. We choose this behavior due to the very simple nature of both the required sensors and actuators needed to theoretically make this behavior emerge. The simple controller used in \cite{FB-MG-RN:21}  had the robots turns clockwise if no other robots are present within the detection region and counterclockwise if at least one robot is present (see Fig.~\ref{fig:milling_figure}).

Understanding the simplicity of the controller, the problem now is how to design, build, and deploy a team of robots that can achieve this task as efficiently as possible. First we must acknowledge the minimum capabilities required of a real robot system to even have a chance of successfully demonstrating the behavior. The minimally viable capabilities required are:

\textbf{Actuation:} move forward while turning either left or right.

\textbf{Sensing:} a binary output indicating whether another robot exists in some forward-facing field-of-view (FOV).

\textbf{Computation:} direct sensing-to-action.

\textbf{Communication:} none.

Note that any robot that is capable of doing at minimum the above is a candidate system for demonstrating the emergent behavior. For instance, even though the behavior is only demonstrated in 2D, this could be done by a swarm of fixed-wing UAVs flying at a fixed height. The remainder of this paper formalizes and applies a novel process to systematically bring the desired milling behavior to life.

The starting point of our problem is the availability of a team of~$N$ homogeneous but idiosyncratic robots (i.e., slightly different from robot to robot) with general unmodeled nonlinear dynamics
\begin{align}\label{eq:realdynamics}
    \dot{x}_i = f(x_i,u_i,\theta^a_i,w_i),
\end{align}
where~$f : \real^d \rightarrow \real^d$ is an unknown function of a robot's state~$x_i \in \real^d$, its control input~$u_i \in \UU \subset \real^m$, the actuation idiosyncrasies~$\theta^a_i \in \real^p$, and the unmodeled disturbances~$w_i \in \real^d$. Depending on the physical sensor(s) used, the robot will need some software able to map the sensor readings to a binary output as required by the binary controller visualized in Fig.~\ref{fig:milling_figure}.

Unfortunately, given this starting point there are no existing works that can help guide exactly how to realize the behavior on the real robots. While many different groups have shown their successful behaviors, it is clear that all the real robot milling behaviors demonstrated so far required a lot of trial and error due to a lack of a way to do standard hypothesis testing. Let us now really make clear the problem at hand that is hidden under the rug in existing published works.

\begin{problem}[Real-world milling with binary sensing]\label{pr:main}
{\rm
    In order to deploy a real world robot swarm experiment one is quickly faced with a lot of choices with no real way of choosing them:
    \begin{enumerate}
        \item How many robots should be deployed?
        \item How fast should they be moving?
        \item How sharp should they be turning?
        \item How good do the sensors need to be?
        \item How homogeneous does the swarm need to be?
    \end{enumerate}
}
\end{problem}

Problem~\ref{pr:main} is not an exhaustive list of questions that need answers and is completely ignoring environmental conditions for instance. It is generally assumed that more robots should always be better in swarms but as this paper shows it is not that simple. Instead, it reveals how complicated searching this space for a successfully emergent behavior can be. If the first experiment deployed doesn't work, how does one know what to modify for the second experiment? Did the experiment fail because the sensors weren't good enough, they were moving too fast, or there were actually \textit{too many} robots on the field? Phase diagrams of simulations have been used by some groups demonstrating the complicated and non-intuitive relationships between the answers to these questions \cite{AC-CKH:18, MRD-YC-ALB-LSC:06, NVB-HA-IYT-SAM:20, ZC-ZC-VT-DC-HZ:16}. 
The goal then is to use low-fidelity simulations-in-the-design-loop to more effectively navigate this space and better understand why experiments fail when they do. This will enable a much more organized and systematic approach to deploying and testing robot swarm systems than is done today.

\section{Methods}\label{se:method}
In this section rather than providing a solution to Problem~\ref{pr:main}, we provide a general simulator-in-the-design-loop framework which can be applied to any real multi-robot system to both (i) search for new emergent behaviors and (ii) bring them to life.

The steps of this `Reality-to-Simulation-to-Reality for Swarms'  (RSRS) process in Fig.~\ref{fig:steps} clearly show the intimate couplings between the simulator (Steps 2 and 3) and the real world robots (Steps 1 and 4). Similar to the work done in \cite{VL-HH-LYC-JW-JI-DS-ML-KG:21, YC-AH-VM-MM-JI-NR-DF:19, LW-RG-QV-YQ-HS-HC:22}, the entire process starts and ends with real robots; however, these works apply their methodologies to train single robots to perform different tasks. We apply our process on a multi-robot system and use the simulator as a tool to help make design choices and guide experiments by enabling hypothesis testing. The steps of the RSRS process are:

\begin{enumerate}
    \item Measure the capabilities/dynamics of robots;
    \item Build simulator;
    \item Exploration within simulation;
    \item Run real robot experiment based on phase diagrams.
\end{enumerate}

\setlength{\parindent}{0cm}\textbf{(1) Measure the capabilities/dynamics of robots.} \setlength{\parindent}{2ex}

The first step of this iterative process is to figure out what aspects of the physical robots need to be simulated. This will vary from system to system (e.g., flying drones, boats, or ground vehicles) but in general this will require the sensing, actuation, computation, and communication abilities of the robots, and ensuring they meet the minimum viability requirements for the given task. In the case of the milling goal from Section~\ref{se:pre}, all we need are robots capable of following a Dubins path with sufficient sensing to generate a binary output signifying the presence of a robot in a particular FOV, but this framework can directly be applied to other tasks/behaviors as well.

Trying to determine the unknown dynamics~$f$ in a parameterized way using~$\theta^a$ in~\eqref{eq:realdynamics} is not the focus of this work. Instead, our starting point is the availability of a simpler kinematic model and mapping~$T : \real^d \rightarrow \real^n$ tracking only a subset of states~$z_i \triangleq T(x_i) \in \real^n$ with~$n \leq d$. Although this may seem like a strong assumption or oversimplification, our goal, again, is not to create a high fidelity simulator but a simpler one that uses only the required information of the robot in order to find emergent behaviors. For example  the 2D milling behavior might be demonstrated using 12-state drones even though we only utilize three (2D projection and orientation) in our simplified simulator.

More specifically we have access to the kinematics
\begin{align}\label{eq:simplifieddynamics}
    \dot{z}_i = g(z_i,u_i,\theta^a_i,w_i) .
\end{align}
Assuming the availability of a function~$g$ parameterized by the actuation idiosyncrasies~$\theta^a_i$, the first task is to take real-world measurements of the actuation/sensing capabilities of the robot in order to synthesize a model that can be used for analysis.

The capabilities of the actuators need to be `measured' by experimentally finding~$\theta^a_i$ for some subset of robots~$i \in \until{N}$. It would not be practical to measure all~$N$ robots, but enough robots should be measured to generate a reliable distribution of the differing~$\theta^a_i$ among the agents.

The capabilities of the sensors/actuators need to be `measured' by experimentally finding how well the hardware can actually be used for their intended purpose. For example in our milling behavior the controller only requires a binary signal but that doesn't necessarily mean a single real-world sensor exists to provide this binary output; some raw signal will need to be processed. The output of our simulated robot~$i$ with binary sensing is given by~$y_i \in \{0, 1 \}$:

\begin{align}\label{eq:realoutput}
    y_i = h(z_i,z_{-i},\theta^s_i),
\end{align}
where~$h : \real^{N\times n} \times \Theta^s \rightarrow \{0, 1\}$ maps the entire simulated world state to a binary output and $\theta^s_i$ characterizes the inaccuracies of the sensor. The robots may have cameras, LIDARs, IRs, or a combination of multiple sensors, however these sensing capabilities can be simplified into a binary, 1-bit signal and the ability at which the real-world system can actually do this is measured and captured in~$h$. 

In terms of computation the controller can essentially be memoryless as all we need is a direct (static) sensing-to-action controller; however, the frequency at which the sensor data can be processed into a binary signal that is fed to an actuator is another measurement that must be taken from the real-world robot. This will provide the timestep~$\timestep$ for the discretized kinematic model used in Step 2.

\setlength{\parindent}{0cm}\textbf{(2) Build Simulator.}\setlength{\parindent}{2ex}

Equipped now with a simplified kinematic model~\eqref{eq:simplifieddynamics}, the distribution of the capabilities of the actuators in the form of a distribution over~$\theta^a_i$, and the sensing capabilities~$\theta^s_i$, we are ready to provide these parameters to our simple simulator.

The second step is to program the information found in Step 1 into the simulator, which should be pretty straightforward depending on the coding language and the previous results. Our simulator is purposely simple enough to make this step as seamless and fast as possible. The outcome of this part of the process should be the discretized (simulated) model of the robot's kinematics and the sensing output,

\begin{align}\label{eq:disc_eqns}
    z_i(\ell+1)  & = z_i(\ell) + g(z_i(\ell),u_i(\ell), \theta^a_i(\ell),w_i(\ell)) \timestep, \notag \\
    y_i(x(\ell)) & = h(z_i(\ell), z_{-i}(\ell), \theta^s_i(\ell)),
\end{align}
where~$z_{-i}$ is the simulated state of all agents except~$i$. 
It is in this step where the disturbances~$w_i$ and idiosyncrasies of the actuation,~$\theta^a_i$, and sensing, $\theta^s_i$ defined in the simulator are purposefully exaggerated to be \textit{worse} than the real-world measurements. Formally, the goal of our simulator~\eqref{eq:disc_eqns} is to have the reachable set of the real trajectories lie within the reachable set of the simulated trajectories,
\begin{align}\label{eq:simulatorreq}
    T(\mathcal{R}(x_i(0),t)) \subset \mathcal{R}(z_i(0),t),
\end{align}
where~$\mathcal{R}(x,T)$ denotes the set of all reachable points within~$T \geq 0$ seconds. In other words, enough noise should be injected into the simulator to include all trajectories the real robot is physically capable of.

Upon completion of Step 2, before moving onto Step 3, it is crucial to ensure (through both simulations and real-world experiments of simple sensing/actuation tests) that~\eqref{eq:simulatorreq} is satisfied. If this is not the case, it may be necessary to return to Step 1 and gather additional information and then reprogram the simulator accordingly until it holds.

\setlength{\parindent}{0cm}\textbf{(3) Exploration within simulation.} \setlength{\parindent}{2ex}

Once the individual robot profiles have properly been measured in Step 1 and built into the simulator and validated in Step 2, the fun part begins. In this step we can leverage decades worth of research dedicated to finding different emergent behaviors such as the works done in the field of Artificial Life~\cite{HS:11,HS:12,AE-JMH:15}. Unlike the vast majority of works that do this type of research, the key difference with our approach is that our simulator was synthesized through real robot measurements, and will ultimately be validated on the real-world robot experiments as well.

This step focuses on trying to discover what the simulated system of agents can do. A popular method of finding how to create certain behaviors is through the use of evolutionary algorithms to optimize a sought after performance metrics as done in \cite{MG-JC-WL-TJD-RG:14, JG-SMO-ALC:18, SAE-JM-OY-KG:18, MG-JC-TJD-RG:14}. In these works, the authors were looking for a predetermined behavior hence why they had a performance metric they wanted to optimize. On the other hand, there is an alternative method where the goal of the algorithm is not to find how to optimize any specific fitness value but rather to find behavioral novelty in the simulation \cite{DB-RT-OH-SL:18, JG-PU-ALC:13, JL-KOS:11}. 
Finding how to create behaviors can also be done through manually changing the different parameters, control algorithms, and/or the environment frequently and observing if anything emerges, although this method is a bit more time intensive. Once any behavior is found, more thorough parameter sweeps can be done with the given control algorithm to see how the changes in parameters affects the behavior; however, only the parameters that can be modified on the real robots should be adjusted.

\begin{figure}[ht]
    \centering
    \includegraphics[width=6.5cm]{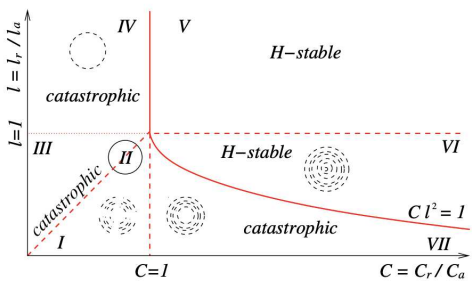}
    \caption{H-stability Phase Diagram of the Morse Potential \cite{MRD-YC-ALB-LSC:06}} \label{fig:phase_diagram2}
\end{figure}

The goal of this step is to allow us to generate a real hypothesis on what we expect to see from the real robot swarm system when deployed. By first ensuring that the emergent behaviors are possible in our simple simulation, the behavior, or phase, of the system can then be identified for each set of parameters and then be plotted in a chart. This diagram will likely contain different phases, or regions, where the different behaviors are formed. Like the phase diagrams of water and other substances \cite{BP-MH-MJP:13}, this diagram will allow us to understand how changes in the parameters affects the system. An example of this is in \cite{MRD-YC-ALB-LSC:06}, where the authors create a phase diagram (shown in Fig.~\ref{fig:phase_diagram2}) that allows them to identify different regions that describe the stability and morphology of the system at different parameter ratios $l = l_r / l_a$ and $C = C_r / C_a$, where $l_r$ and $l_a$ represent the repulsive and attractive potential ranges, and $C_r$ and $C_a$ represent their respective amplitudes. 

\setlength{\parindent}{0cm} \textbf{(4) Run real robot experiment.} \setlength{\parindent}{2ex}

The results from the exploration step can be used to set up the controls and conditions for experiments using the real system of robots. Just like a phase diagram can be used to determine the phase of water in different temperature/pressure conditions \cite{BP-MH-MJP:13}, one can use something like a phase diagram shown in Fig.~\ref{fig:phase_diagram2} to make hypotheses about real world swarm deployments and help guide sequential experiments.

Of course the goal is for the real-world robots to immediately exhibit the same behavior we find through simulation given the same parameters and conditions, this naturally will not always be the case. Or more troublesome, given our approach shown in Fig.~\ref{fig:gap} of acknowledging a large reality gap, the simulator may not be very predictive of the real world experiment in general.

Given a particular world setup (either in simulator or the real world), we let $B_r(\cdot)$ and~$B_s(\cdot) \in [0,1]$ be the unknown probability of the conditions leading to the desired emergent behavior (in this case milling) if given enough time. Lumping all aspects of the simulator into a variable~$E$ (e.g., number of robots, type of controller used, sensor quality, environmental disturbances) and hope for the relation that if something is discovered in the simulator, there is a higher probability of it occurring in the real world experiment than not,
\begin{align}\label{eq:finalgap}
    B_r(E) \geq B_s(E).
\end{align}

By exploring in the simulator for conditions~$E$ for which~$B_s(E) \approx 1$, we can be more confident that the real world system will also demonstrate the same emergent behavior.

There is also the possibility that the simulation in Step 3 isn't producing any behavior in the system. In other words, no matter what kinds of simulated environments and conditions~$E$ are tried, the value~$B_s(E)$ cannot be made close to 1 meaning the capabilities of the simulated robots are not good enough to manifest the behavior seen in a more capable robot.
It is here where the cost of continuing this method with the given capabilities should be analyzed. If there are still simple things that can be done to modify the simulator that wouldn't take much time or effort (like reducing the amount of added noise) then that should be the first course of action. However, if a behavior still isn't produced and the modifications to the simulator would be far too time-consuming, then it may be concluded that the robots are at that point simply too incapable to create any behavior.  Small upgrades and changes to the robot system should then be done and the RSRS process should start over with Step 1 since the capabilities of the robots have now changed.

\section{Case Study }\label{se:case}
Here we show how our general framework proposed in Section~\ref{se:method} can be applied to a real robot system attempting to recreate an emergent behavior observed only in simulation.

\begin{figure}[ht]
    \centering
    \subfigure[]{\includegraphics[width=.25\linewidth]{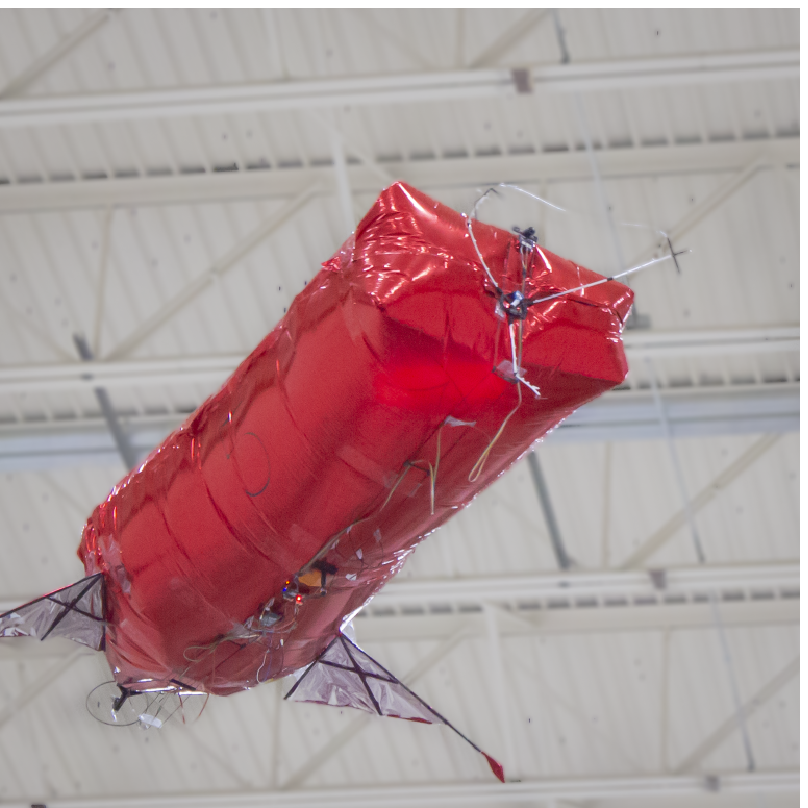}}
    \subfigure[]{\includegraphics[width=.32\linewidth]{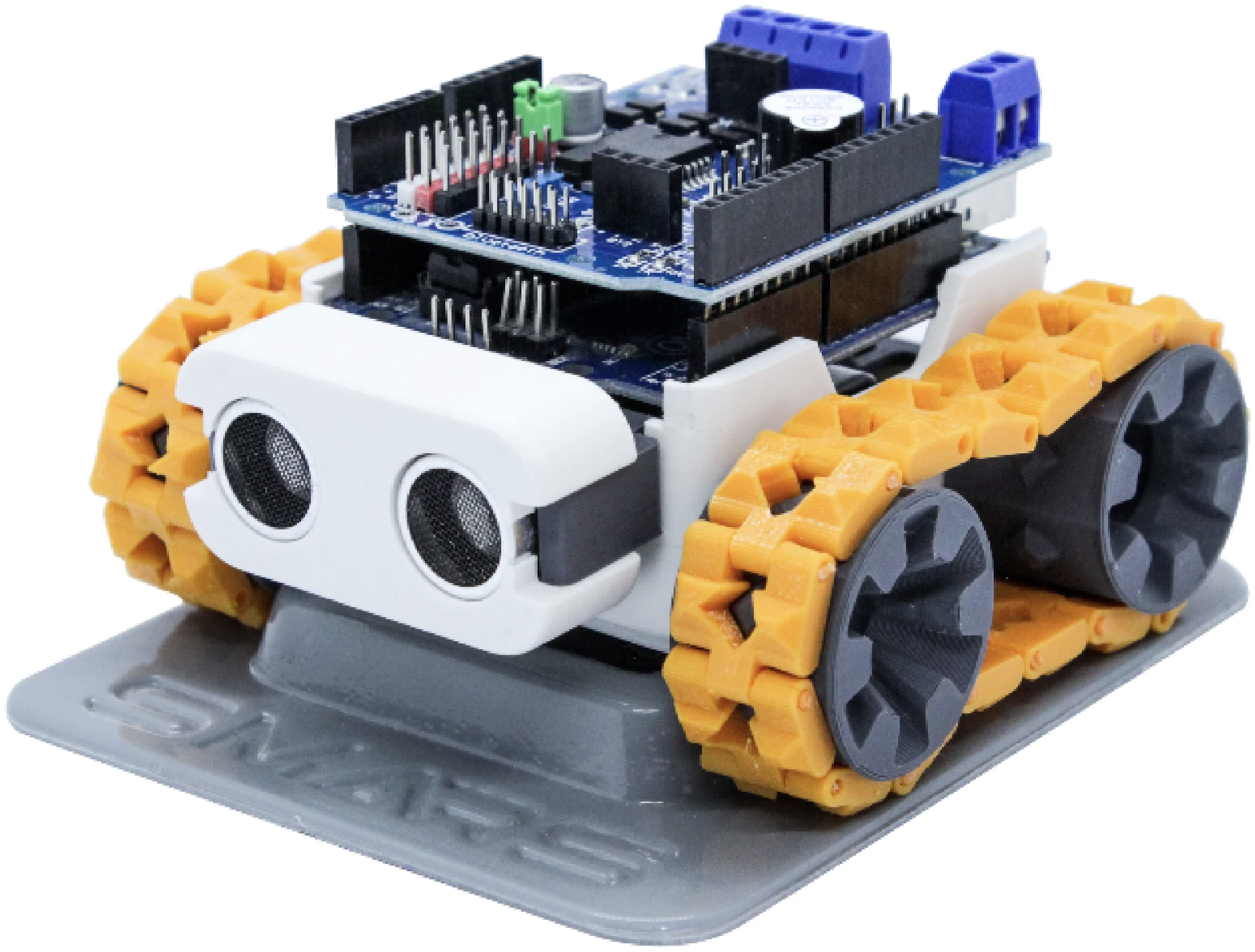}}
    \subfigure[]{\includegraphics[width=.25\linewidth]{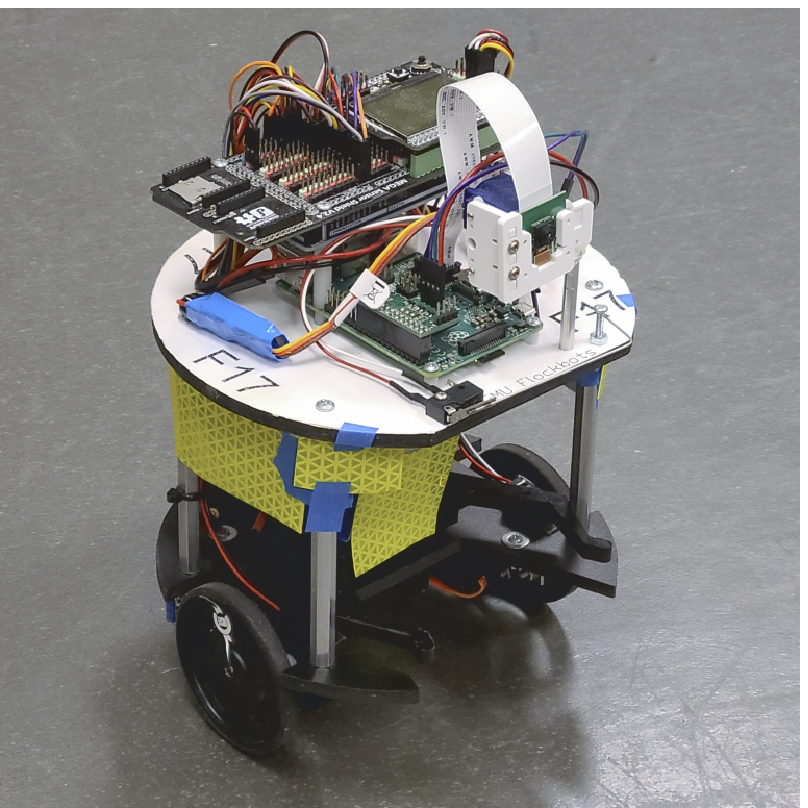}}
    \caption{Images of real robot platforms built for different purposes used in various labs or a makerspace showing (a) SPARX lighter-than-air robots (b) SMARS robots, and (c) FlockBots.}\label{fig:flockbots}
\end{figure}

The main robot platform used were the Flockbots shown in Fig.~\ref{fig:flockbots}(c), which were designed and built years prior to this research. The Flockbots are two-wheeled differential drive robots that are 15cm in diameter. They are equipped with several sensors including wheel encoders, IR sensors, an RGB camera, and more connected to a Raspberry Pi computer. The wheel speed is governed by an Arduino microcontroller which maintains the robot’s forward speed and turning rate in closed-loop control. Our only modifications to the system are the control logic, which uses only a binary output from the center forward-facing IR sensor to turn left or right respectively at a constant speed, and retro-reflective tape on the perimeter of each robot to increase the inter-robot detection range. The IR sensor generates the output as 1 when the distance measured is less than the maximum value of the sensor, otherwise the binary output is a 0.

Following the steps of Section~\ref{se:method}:

\noindent \textbf{(1) Measure the capabilities/dynamics of robots.}

These robots use differential drive to operate, so by moving the wheel on one side at a different speed than the other,  the robots are able to turn their heading and move at different speeds. We then modeled the simulation of these robots as a unicycle model with the kinematics shown in \eqref{eq:real_func}. 
In this case the reduced order model is given by~$n = 3$ states for each robot~$z_{i,1}$ and~$z_{i,2}$ representing the 2D position and~$z_{i,3}$ representing the orientation of robot~$i$, respectively. Since these robots are ground vehicles with no slipping, we are able to omit~$w_i$ and capture differences in the robots' capabilities through the idiosyncrasies~$\theta^a_i$. 

\begin{align}\label{eq:real_func}
    g  = \left[ \begin{array}{c} u_{i,1} \theta^a_{i,1} \cos z_{i,3} \\ u_{i,1} \theta^a_{i,1} \sin z_{i,3} \\ u_{i,2} \theta^a_{i,2} \end{array} \right]
\end{align}

\textbf{Actuation:}
The first step of this process is to measure the actuation capabilities of the robots. We begin by determining the maximum and minimum safe control inputs. Then, we gave the Flockbots constant control inputs at intervals within the safe bounds and calculated the average speed from the time taken to cross a 2 meter track. We performed multiple trials for each speed across a representative group of five Flockbots, a sample of these measurements for three different robots can be seen below in Table~\ref{tab:speed1}. From these measurements, we found the mean and standard deviation for the measured speeds of each robot at each set value. The $\theta^a_1$ values of each tested robot was found by using the following equation:

\begin{align}\label{eq:theta^a1}
    \theta^a_1 = \frac{\text{Average~Measured~Speed}} {\text{Desired~Speed}}
\end{align}

\begin{table}[h]
    \begin{center}
        \begin{tabular}{|l|l|l|l}
            \hline
            Desired Speed (mm/s) & Error Distribution            & $\theta^a_1$ \\
            \hline
            $u_1 = 25$           & $v_1$ $\sim$ $N(22.15, 0.57)$ & 0.89       \\
            $u_1 = 25$           & $v_2$ $\sim$ $N(21.48, 0.04)$ & 0.86       \\
            $u_1 = 25$           & $v_3$ $\sim$ $N(22.32, 0.46)$ & 0.89       \\
            $u_1 = 50$           & $v_1$ $\sim$ $N(48.47, 0.38)$ & 0.97       \\
            $u_1 = 50$           & $v_2$ $\sim$ $N(48.66, 0.12)$ & 0.97       \\
            $u_1 = 50$           & $v_3$ $\sim$ $N(48.54, 0.20)$ & 0.97       \\
            \hline
        \end{tabular}
    \end{center}
    \caption{Speed Measurements}
    \label{tab:speed1}
\end{table}

The measurements for the turning rates were conducted in a similar manner. The robots were programmed to turn in place at constant rates and the actual turning rate was measured by allowing the robots to turn in a complete circle and dividing by the time they took to do so. The values for $\theta^a_2$ were found using a similar equation to how $\theta^a_1$ was calculated:

\begin{align}\label{eq:theta^a2}
    \theta^a_2 = \frac{\text{Average~Measured~Turning~Rate}} {\text{Desired~Turning~Rate}}
\end{align}

\begin{figure}[ht]
    \centering
    \includegraphics[width=6cm]{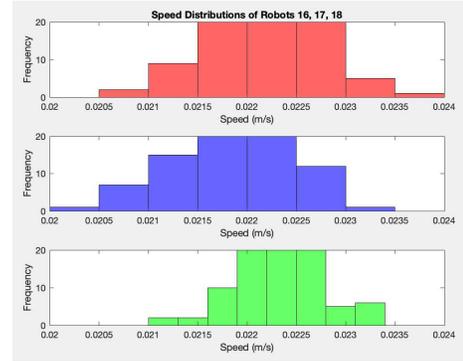}
    \caption{Measured Speed Distributions of Flockbots}
    \label{fig:distribution}
\end{figure}

By finding these distributions for the individual robots, we can predict the overall distribution of all measured robots to find how much the speeds and turning rate can vary in the system. Of course this data doesn't encompass all the information of the whole system but it gives us an idea of how inaccurate and idiosyncratic the robots may be. These idiosyncrasies of the Flockbots can be seen in the several different gaussian distributions of the measured means and standard deviations of the robot speeds in Fig.~\ref{fig:distribution}.

\begin{figure}[ht]
    \centering
    \includegraphics[width=7cm]{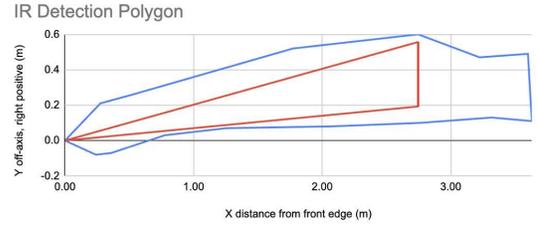}
    \caption{Measured detection range of Flockbot in blue and simulated detection range in red }
    \label{fig:sim_sensing}
\end{figure}

\textbf{Sensing:}
Additionally, tests were also done to find the accuracy of the IR sensors on board the robots. We gathered data on the detection accuracy with which a single Flockbot could detect another Flockbot at different distances and angles. Fig.~\ref{fig:sim_sensing} shows the 80$\%$ positive detection threshold as the blue polygon. The IR sensor detected another Flockbot on this line at least 80$\%$ of the time, with nearly 100$\%$ positive detection for Flockbots within this region thanks to the retro-reflective tape. We also measured a 5-8$\%$ false positive rate with no other Flockbots nearby.

\textbf{Computation:} From our tests, we found that the robots' sampling period were around 130ms. This value was used as the time step value ($\timestep$) for \eqref{eq:disc_eqns}.

\textbf{Communication:} None.

\noindent \textbf{(2) Build simulator.}

For the sake of using an extremely intuitive and easy-to-modify simulator, we utilize NetLogo \cite{UW:99} to rapidly simulate the measured capabilities of the robots under different conditions and in different environments using~\eqref{eq:disc_eqns} with the measured parameters~$\theta^a_i$. 

\textbf{Actuation:}
The size of the robots were programmed to directly map to the collision radius of each agent, if an agent's position is within double the collision radius, then the simulator would consider it a collision. When there was a collision with the real Flockbots, the robots would stop in place if the other robot was in front and would be generally unaffected if it was pushed from another angle, all of which was programmed into NetLogo.

The speed was set to be how many patches (standard block on NetLogo, set to represent 0.1m) the agents would move per second. The turning rate was how many radians the heading of the agent would change per second.

For the noise/disturbances, we first found the distribution of how far off the agents would move from the target in the real world experiments and the reliability distribution of the IR sensor as mentioned in Step (1). Once we had a sample of the real distribution of noise, we expanded it such that the agents in the simulation would obtain their $\theta^a$ values from the normal distribution found from all the measured $\theta^a$ values. For example, when the desired speed was set to be 25 mm/s, the robots all had slightly different true speeds, so the agents in simulation were set to have a value within a larger distribution.  This made the agents in simulation more idiosyncratic/random and less reliable than their real world counterparts, thus making the agents in the simulation worse than the real robots.

The discretized equations for the kinematics and the output were then modeled to be:

\begin{align}\label{eq:disc_func}
g(\ell)  = \left[ \begin{array}{c} u_{i,1}(\ell) \theta^a_{i,1}(\ell) \cos z_{i,3}(\ell)\\ u_{i,1}(\ell) \theta^a_{i,1}(\ell) \sin z_{i,3}(\ell)\\ u_{i,2}(\ell) \theta^a_{i,2}(\ell)\end{array} \right]
\end{align}

\begin{align}\label{eq:disc_kine}
z_i(\ell + 1)  = \left[ \begin{array}{c}
z_{i,1}(\ell+1) \\ z_{i,2}(\ell+1) \\ z_{i,3}(\ell+1) \end{array} \right] = \left[ \begin{array}{c} z_{i,1}(\ell) + g_1(\ell)\timestep \\  z_{i,2}(\ell) + g_2(\ell)\timestep \\ z_{i,3}(\ell) + g_3(\ell) \timestep \end{array} \right]
\end{align}

\textbf{Sensing:}
From the information gathered, we can model the sensing function of the robots as the function:
{
\begin{align}\label{eq:output}
    h_i(z) = \begin{cases}
                 1 & \text{if } \exists j \neq i, s.t. ~z_j \in \operatorname{FOV}_i \\
                 0 & \text{otherwise.}
             \end{cases}
\end{align}
}

where the FOV is the conical area in front of the sensor with a measured distance and opening angle.

The agents' FOV was originally believed to be made up by the vision-distance and vision-cone values (which when combined gives a conical sensing region in front of the agents). However, from the measurements conducted, we now know that there are areas in the real sensors that are more reliable than other areas, making it more of a detection polygon rather than a cone. Despite the rough qualities of the real sensor shown in blue in Fig.~\ref{fig:sim_sensing}, we program a simpler polygon into our simulator as shown in red. This again ensures that our simulated agents are simple but less capable than the physical robots as our approach shown in Fig.~\ref{fig:gap}(c) shows.

\textbf{Computation:} Simulator is set to run at the same frequency $\timestep$ as the real robots.

\textbf{Communication:} None.

To move on to Step 3, it was verified that~\eqref{eq:simulatorreq} was satisfied for all conditions considered in both simulation and real world experiments.

\noindent \textbf{(3) Exploration within simulation.}

Since our goal was to recreate milling, here we started with using the control algorithm similar to ones that were used in \cite{FB-MG-RN:21} and \cite{DS-CP-GB:18} to make their robots mill (shown in Fig.~\ref{fig:milling_figure}), 
\begin{align}\label{eq:controller}
u_{i,1}(\ell) &= v,  \\
u_{i,2}(\ell) &= \begin{cases} \omega, \quad \text{if } y_i = 1 , \\ -\omega, \quad \text{otherwise.} \end{cases} 
\end{align}
where $v$ is the desired forward speed and $\omega$ is the desired turning rate programmed into the robots.

Admittedly, since we already knew exactly what we were searching for we did not require too much genuine `exploration'. 
We found that the milling behavior was produced in our simulations using the classic control algorithm mentioned prior, but only in certain cases. The parameters that were changed were only ones that we could control (e.g., desired speed, number of robots, desired turning-rate). The parameters that weren't changed were the vision-cone and vision-distance of the IR sensor, this is because the sensor on the robots were fixed and was used as a binary sensor that was set to only output a detection when it had any non-zero readings. The initial positions of the agents in simulation also had a large effect on the outcome of the system, if they were set too close to each other, the possibility of collisions increased dramatically. On the other hand, if they were randomly positioned in the environment, there was a high likelihood that some of the agents would never see any other agent and permanently stay moving within their own circular path. Therefore, we initialized the agents to be randomly positioned within a circle with radius slightly larger than their maximum vision distance and to also be facing directly away from the center of the environment; this minimized collisions and also allowed the agents to interact with others at least once.

\begin{figure}[ht]
    \centering
     \subfigure[]{\includegraphics[width=.99\linewidth]{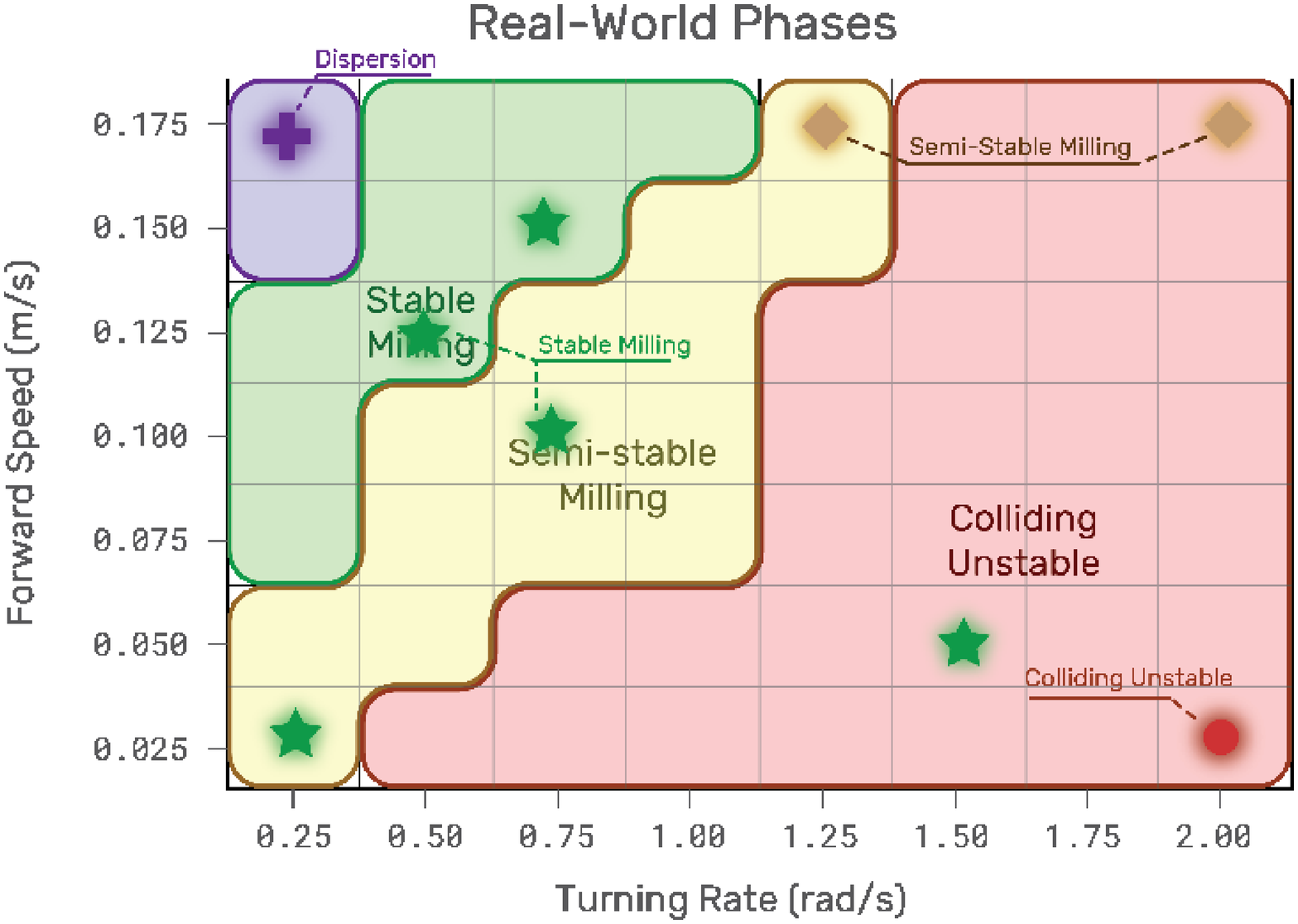}}
    \caption{A 2D slice of a minimally viable phase diagram  found in simulation overlaid with markings showing nine real experiments run out of which 5 cases has a exact match in simulation and reality whereas in the remaining 4 cases the real robots had an easier time milling as desired in~\eqref{eq:finalgap}.}
    \label{fig:phase_plot2}
\end{figure}

Then we created multiple plots to visualize how these changes affected the quality of the milling circle produced. The phase at each point was determined by the researchers; and although it may be subjective, the plots give a useful visualization of how the parameters affect the system. An example of these phase diagrams can be seen for~$N=9$ agents in Fig.~\ref{fig:phase_plot2}(a). It should be noted that we are essentially viewing a 2D slice of a much higher dimensional space in which the `phase' of the entire swarm can entirely change based on a single parameter/condition being different. The phase diagram contains four regions: 
\newline \textit{Dispersion} represents a behavior when the agents weren't able to form any shape and instead spread out towards the walls of the environment.
\newline \textit{Stable Milling} is the behavior that most closely represents a circle (i.e. the milling region).
\newline \textit{Semi-stable Milling} would continuously be colliding with each other though the group would still be rotating around a center.
\newline \textit{Colliding Unstable} is where the agents collide in a manner that prevented them from continuing to move.

From these plots we found that, in the simulator, the set of conditions/parameters with the better chances of producing a circle was with 9 agents positioned to face away from the center of the environment and set to move at a speed of 0.15 m/s and a turning rate of 0.75 radians per second, since it was the point in the center of the circle phase region. While this initial process is not guaranteed to work on the real-world experiment, it provides a much better set of initial answers to Problem~\ref{pr:main} than is available today for a never-before-deployed robot swarm. 

\noindent \textbf{(4) Run real robot experiment}

\begin{figure}[ht]
    \centering
    \subfigure[]{\includegraphics[width=.4\linewidth]{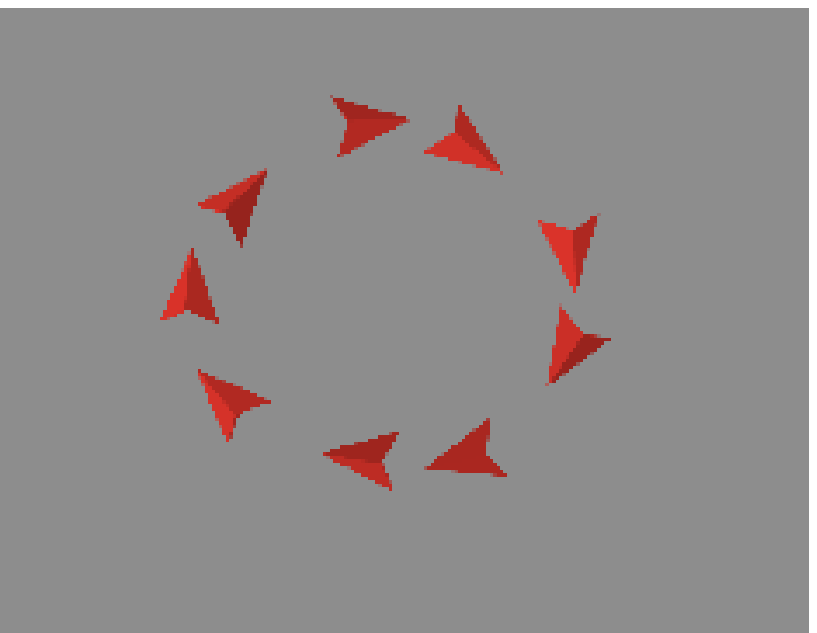}}
    \subfigure[]{\includegraphics[width=.4\linewidth]{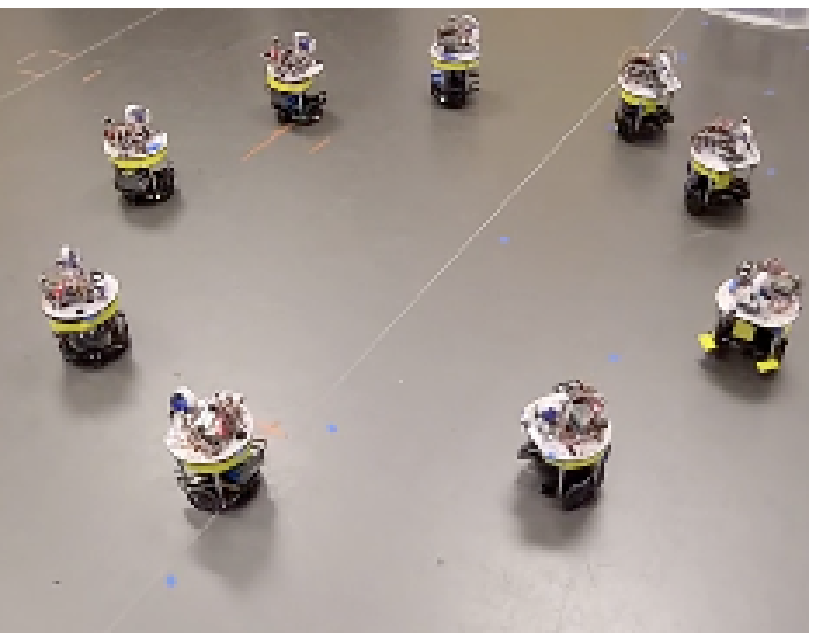}}
    \caption{(a) Nine agents milling in NetLogo simulator and (b) nine Flockbots successfully milling using the same parameters.}
    \label{fig:milling}
\end{figure}

We then attempted to run an experiment on the robots using this recommendation that, according to our simulations, has the highest chance of producing the found behavior. Remarkably, the 9 real robots were able to successfully mill at the given parameters the very first time we ran an experiment involving all 9 robots. The simulated agents can be seen in Fig.~\ref{fig:milling}(a) and the real robot experiment is shown in Fig.~\ref{fig:milling}(b). The main novelty in this approach is a method of creating minimally viable phase diagrams to help make informed decisions on how to deploy a real-world robot swarm. 

We then ran multiple experiments after the successful first attempt to see how well the simulated phase diagrams predicted the behaviors of the real robot system at different sets of parameters. We documented the behavior of the real robot system and created the diagram shown in Fig.~\ref{fig:phase_plot2} showing the results of nine real experiments with the tested parameters overlaid over the phase diagram obtained from simulations in Step~3.

As seen, the real robots were able to mill at a speed of 0.15 m/s and a turning rate of 0.75 radians/s, just as the simulated phase diagram predicted. Additionally, the robots recreated this milling behavior at different points where the simulated agents weren't able to mill. This shows that~\eqref{eq:finalgap} seemed to hold true for our admittedly few number of samples. However, the fact that we were able to not only create the milling behavior successfully on the first full scale experiment and have relative high reliability for slight changes in parameters was enabled due to the systematic RSRS process. Without this simulation-in-the-loop process, it certainly would have taken us a lot more tinkering to find successful milling combinations.

It should be noted that before finding success using the Flockbots shown in Fig.~\ref{fig:flockbots}(c), we first applied our methodology to Lighter-Than-Air (LTA) robots shown in Fig.~\ref{fig:flockbots}(a) and discovered after Step 2 that the real-world disturbances were far too large such that our simulated world with even harsher conditions didn't exhibit any desirable behaviors. We then applied the RSRS process to SMARS robots shown in Fig.~\ref{fig:flockbots}(b) and discovered that the sensors used on them were so bad that when made worse in simulation, the simulated agents were not capable of anything either. While discouraging, these initial simple simulations and measurements were sufficient in justifying why upgrading the physical robots was even necessary. Without this process, we may have spent several months tinkering with either of the prior sets of robots without even realizing we never stood a chance of replicating the desired milling behavior.

\section{Conclusions}\label{se:conclusions}
In this paper we challenge the immediate desire to bridge the simulation-reality gap, particularly when considering multi-robot swarm systems. In this work we have only focused on a single emergent behavior (circular milling) that is achievable by the robots using only a binary sensor. However, the framework proposed is applicable to any real homogeneous (but idiosyncratic) multi-robot swarm for which collective/swarm behavior might be a possibility. More specifically, we show how even simple simulators can be used to help guide multi-robot control designs for even more complicated systems without necessarily shrinking or `bridging the gap' between simulation and real robots. The primary contribution is a simulation-in-the-design-loop framework for robot swarm systems that can be used as a tool to not only help understand the connections between local interactions and globally emergent behaviors but also bring them to life in real multi-robot systems.

\section*{Acknowledgments}
This work was supported in part by the Department of the Navy, Office of Naval Research (ONR), under federal grants N00014-20-1-2507 and N00014-22-1-2207. 

\bibliographystyle{ieeetr}
\bibliography{ricardo}

\end{document}